\documentclass[sigconf]{acmart}
\usepackage{array}
\usepackage{booktabs}
\usepackage{multirow}
\usepackage{colortbl}
\usepackage{tabularx}
\usepackage{tikz}
\definecolor{bestpink}{RGB}{255,225,234}
\definecolor{secondblue}{RGB}{225,240,255}
\definecolor{caskgblue}{RGB}{230,242,255}

\newcommand{\bestresult}[1]{\textbf{#1}}
\newcommand{\secondresult}[1]{\underline{#1}}
\usetikzlibrary{arrows.meta,positioning}

\AtBeginDocument{%
  }

\setcopyright{none}
\copyrightyear{2026}
\acmYear{2026}
\acmDOI{}
\acmISBN{}

\title[CaSKG]{CaSKG: Counterfactual-Causal Skill Graphs for Scalable Agent Skill Retrieval}

\author{Zhiyuan Li}
\authornote{Equal contribution; work done while Zhiyuan Li was an intern at Ant Group.}
\affiliation{%
  \institution{School of Artificial Intelligence,\\ Jilin University}
  \institution{Ant Group}
  \city{Changchun}
  \country{China}
}
\email{zhiyuanl24@mails.jlu.edu.cn}

\author{Linyuan Gao}
\authornotemark[1]
\affiliation{%
  \institution{School of Artificial Intelligence,\\ Jilin University}
  \city{Changchun}
  \country{China}
}
\email{lygao25@mails.jlu.edu.cn}

\author{Xuechun Ding}
\affiliation{%
  \institution{Ant Group}
  \city{Hangzhou}
  \country{China}
}
\email{dingxuechun.dxc@antgroup.com}

\author{Hongwei Chen}
\authornote{Co-corresponding authors.}
\affiliation{%
  \institution{Ant Group}
  \city{Hangzhou}
  \country{China}
}
\email{wei.chenhw@antgroup.com}

\author{Yuan Wu}
\authornotemark[2]
\affiliation{%
  \institution{School of Artificial Intelligence,\\ Jilin University}
  \city{Changchun}
  \country{China}
}
\email{yuanwu@jlu.edu.cn}

\author{Yi Chang}
\affiliation{%
  \institution{School of Artificial Intelligence, \\ Jilin University}
  \city{Changchun}
  \country{China}
}
\email{yichang@jlu.edu.cn}

\renewcommand{\shortauthors}{Z. Li, etal.}

\begin{document}

\begin{abstract}
Reusable skill libraries allow large language model (LLM) agents to reuse procedural knowledge across tasks, but they also turn memory access into a challenging retrieval problem. Full-library prompting preserves coverage at high context cost, vector retrieval returns compact neighborhoods but treats skills as independent text, and graph-based retrieval can recover workflow context only when the edges that carry relevance are reliable. We propose CaSKG, a counterfactual-causal skill graph framework that calibrates procedural relations before retrieval. CaSKG first builds a high-recall directed candidate graph from semantic, lexical, input/output, and structural evidence, with repair evidence and an optional LLM judge further refining candidate scores. It then applies direction-conditioned textual counterfactual probes that remove, substitute, and reorder skill pairs, aggregates the evidence with Bayesian smoothing, and publishes a state-filtered weighted graph for task-conditioned expansion. The graph is constructed offline and used without changing the downstream agent policy or task interface. Across six LLM backbones on ALFWorld ID-140 and ScienceWorld U211, CaSKG achieves the highest task score in all twelve combinations of model and benchmark. Relative to Graph-of-Skills (GoS), it improves the six-model macro-average ScienceWorld score from 72.62 to 80.50 and ALFWorld success from 80.01\% to 86.79\%, while reducing mean environment steps on both benchmarks. Qualitative and ablation analyses further show that calibrated edges help retrieval preserve prerequisites, state-changing actions, verification routines, and final completion steps. These results position edge-confidence calibration as an effective route to compact and executable skill retrieval at scale\footnote{Code is available at: \url{https://github.com/ZhiyuanLi218/Caskg}}.
\end{abstract}



\maketitle

\section{Introduction}

Large language model (LLM) agents increasingly solve tasks by combining generation with external tools and application programming interfaces (APIs), and interacting with environments. Toolformer shows that language models can learn to invoke APIs during generation~\cite{schick2023toolformer}, and ReAct interleaves reasoning with actions in an environment~\cite{yao2023react}. Beyond one-time tool calls, agents can also accumulate reusable procedures: Voyager, for example, stores executable skills that can be retrieved and reused across tasks~\cite{wang2024voyager}. Augmented language models and systems built for large tool repositories further enlarge the action space available to an agent~\cite{mialon2023augmented,patil2024gorilla,qin2024toolllm}. As this procedural memory grows, the central question is no longer whether an agent can use a tool, but how it can expose the right subset of skills for the current task without overwhelming the context.

Skill retrieval is difficult because useful task context is rarely a single textually matching procedure. A household or science task may require prerequisites, object-location routines, state-changing actions, verification steps, and recovery procedures whose descriptions do not all overlap with the task instruction. Full-library exposure preserves recall by making every skill visible, but it shifts the selection burden to the agent and introduces many irrelevant alternatives. Dense retrieval and retrieval-augmented generation (RAG) reduce the context size by ranking items according to semantic similarity~\cite{karpukhin2020dpr,lewis2020rag}, yet they usually treat skills as independent text units. This is a poor fit for procedural memory, where the utility of one skill often depends on another skill that prepares, checks, repairs, or follows it. Recent tool-retrieval evidence reinforces this mismatch: textual retrieval scores alone do not reliably predict whether a returned capability will be useful for the current task~\cite{shi2025toolret}.

Graph-based skill retrieval is a natural response to this limitation. Instead of retrieving isolated skills, GoS constructs an offline skill graph and propagates query relevance from lexical and semantic seeds so that workflow-related skills can enter the retrieved bundle~\cite{liu2026graph}. This relational view addresses an important weakness of vector-only retrieval: a skill can be operationally necessary even when it is not the closest textual neighbor of the task instruction. However, graph retrieval introduces a different failure mode. Once relevance starts to propagate, the quality of the retrieved bundle depends on the edges that carry that relevance. An edge induced by topical similarity, co-occurrence, or loose interface compatibility may look plausible during construction but still connect skills that are alternatives, unordered neighbors, or operationally unsuitable for the current procedure.

The bottleneck is therefore not simply how to build a larger skill graph, but how to decide which candidate relations should influence retrieval. Publishing every plausible association increases coverage but can spread relevance through weak edges and pollute the skill context. Pruning aggressively can remove useful paths and leave parts of the library unreachable. Exhaustively assessing all ordered skill pairs is also impractical because the number of possible relations grows quadratically with the library size. This paper formulates skill-graph construction as a budgeted edge-confidence calibration problem: given a high-recall pool of candidate relations, the system must decide which edges deserve assessment, how much confidence each assessed relation should receive, and how that confidence should control graph propagation.

We propose CaSKG, a counterfactual-causal skill graph framework that separates association discovery from edge reliability assessment. CaSKG first induces a high-recall directed candidate graph from skill-level evidence, including semantic, lexical, input/output, and structural signals, with repair evidence and an optional LLM judge refining candidate scores. The framework also reserves trace co-occurrence and existing-relation channels for self-evolving skill libraries; in the static construction used in this study, these channels remain extension points rather than active inputs. The active construction signals are used to discover plausible source-to-target hypotheses rather than to directly publish every relation at full strength. CaSKG then allocates a limited validation budget to selected candidate edges and evaluates each directed hypothesis with three textual counterfactual probes: removing the source skill, replacing it with a dissimilar skill, and reversing the proposed order. These probes estimate whether the target skill depends on the source, whether the source is specific rather than interchangeable, and whether the relation has a meaningful workflow direction~\cite{li2023counterfactual,nguyen2024counterfactuals}.

The probe outputs are aggregated with a Beta-smoothed posterior that yields an edge-level reliability estimate. CaSKG converts this estimate into a publication state: confirmed relations retain full support, uncertain relations are downweighted, rejected relations are removed, and a bounded set of unvalidated candidates is kept only as a low-weight scaffold for coverage. The resulting graph is frozen before evaluation. At runtime, lexical and semantic matches initialize a personalized PageRank-style expansion over this state-filtered graph, so the downstream agent receives a compact skill bundle shaped by calibrated procedural structure rather than by raw association strength alone. CaSKG therefore changes the graph over which retrieval propagates without adding a new online action policy or changing the task interface.

We evaluate CaSKG against full-library exposure, vector retrieval, and GoS on ALFWorld ID-140 and ScienceWorld U211 with six LLM backbones~\cite{shridhar2020alfworld,wang2022scienceworld}. In the complete archived comparison, CaSKG obtains the highest reported task score in all 12 backbone and benchmark groups. Relative to GoS, the six-model macro-average ScienceWorld score rises from 72.62 to 80.50, and ALFWorld success rises from 80.01\% to 86.79\%. Additional task-type, trajectory, scale, and component analyses show that the gains are consistent with the intended mechanism: improving the reliability of graph edges helps retrieve skill bundles that preserve prerequisites, intermediate state changes, verification routines, and final task-completion actions.

The main contributions of this work are:
\begin{itemize}
    \item We formulate scalable skill-graph construction as a budgeted edge-confidence calibration problem, where retrieval must balance relational coverage against the risk that weak associations distort graph propagation.
    \item We develop CaSKG, an offline graph-construction method that combines multi-signal candidate induction with LLM-simulated counterfactual probes and Bayesian state-gated publication before runtime retrieval.
    \item We evaluate CaSKG on two interactive benchmarks with six LLM backbones and analyze its behavior through aggregate results, task-type gains, trajectory examples, library-scale records, and graph-construction ablations.
\end{itemize}

\begin{figure*}[t]
  \centering
  \includegraphics[width=\textwidth]{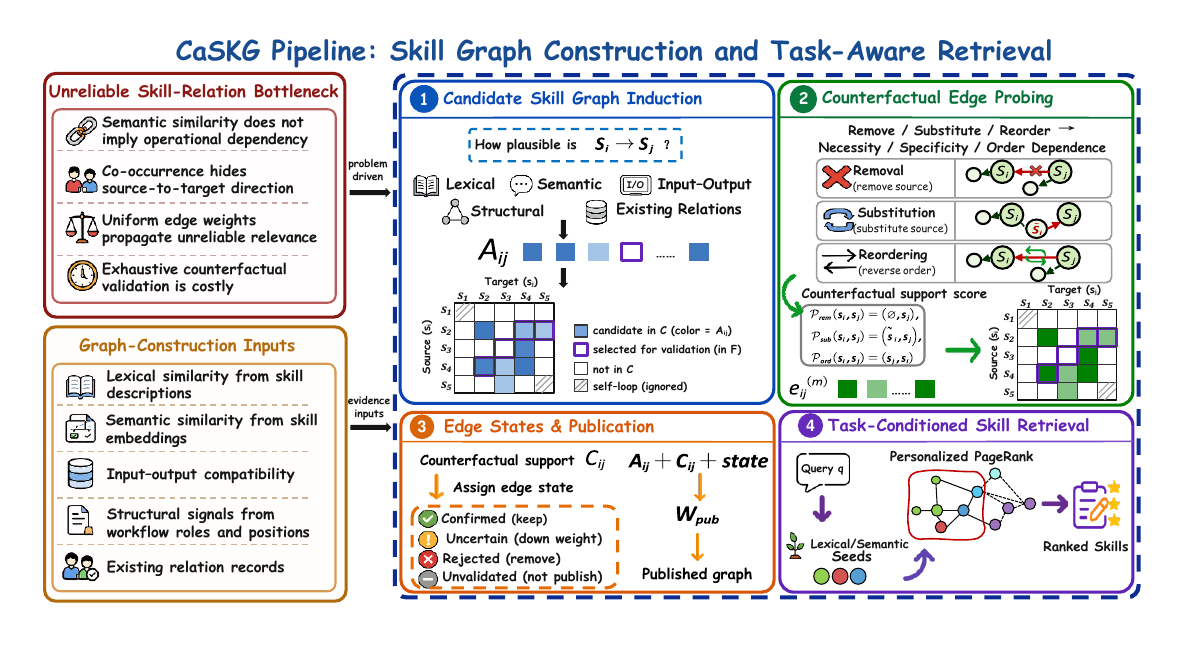}
  \caption{Overview of CaSKG. Multiple skill-level signals induce a directed candidate set. A budgeted subset is examined with direction-conditioned removal, substitution, and reordering probes. The resulting evidence assigns edge states and publication weights, producing the published skill graph. Task-conditioned retrieval then propagates query relevance over this graph to return a compact skill context.}
  \Description{A four-stage CaSKG pipeline. Candidate relations are induced
from skill-level signals, selected edges are evaluated with removal,
substitution, and reordering probes, edge states determine graph publication,
and personalized PageRank retrieves task-relevant skills from the published graph.}
  \label{fig:caskg-pipeline}
\end{figure*}

\section{Related Work}

\paragraph{Tool retrieval.}
Tool-augmented language models have evolved from local invocation decisions toward retrieval over large and reusable capability libraries. Early work framed tool use as a control problem inside generation: Toolformer learns to insert API calls into model outputs, ReAct couples reasoning traces with environment actions, and augmented-language-model surveys place these behaviors in a broader family of models that call external modules and environments~\cite{schick2023toolformer,yao2023react,mialon2023augmented}. As tool repositories grow, the challenge is no longer limited to deciding whether a call should be made. Gorilla connects language models to large API collections, while API-Bank, ToolBench, and ToolLLM turn large-scale tool selection, planning, and composition into explicit evaluation settings~\cite{patil2024gorilla,li2023apibank,xu2023toolbench,qin2024toolllm}. A further step is to treat reusable action sequences as skills rather than isolated interfaces: Voyager stores executable skills across tasks, and ToolRet shows that generic retrieval scores are not reliable indicators of whether a returned tool will solve the current task~\cite{wang2024voyager,shi2025toolret}. This line of work establishes the need for retrieval, but it leaves open how an agent should retrieve a coherent set of interdependent procedures rather than a single callable tool.

\paragraph{Graph retrieval.}
The need to retrieve interdependent capabilities connects skill retrieval to a broader movement from independent matching toward structured memory. Dense Passage Retrieval and RAG provide scalable query-to-memory retrieval, but their standard use largely scores each unit independently~\cite{karpukhin2020dpr,lewis2020rag}. Graph-based retrieval adds a second source of evidence: relations among memory units. Topic-Sensitive PageRank shows that query-biased propagation can use links among candidates, and recent systems such as GraphRAG and HippoRAG adapt graph-structured memory to corpus and knowledge retrieval~\cite{haveliwala2002topicsensitive,edge2024graphrag,gutierrez2024hipporag}. This idea has also moved into tool and skill settings. ToolNet represents relations among tools, Graph-of-Skills builds a dependency-aware graph for executable skills and retrieves bounded skill bundles through graph propagation, GraSP models skill composition with typed preconditions and effects, and SkillReranker uses task states to refine the candidate order~\cite{liu2024toolnet,liu2026graph,xia2026grasp,chen2026task}. These methods show that relational structure can recover useful capabilities missed by one-shot semantic similarity. Their effectiveness, however, depends on the reliability of the relations over which relevance propagates. Edges induced from semantic similarity, co-use, interface compatibility, or workflow order may be plausible associations without being valid procedural dependencies. Once such edges are published, multi-hop propagation can carry relevance toward unrelated or operationally unsuitable skills. The assessment of candidate relations before they influence skill retrieval therefore remains a central unresolved issue.

\paragraph{Causal graph retrieval.}
Causal graphs provide a complementary perspective on relation quality because they distinguish directional dependence from statistical association. The potential-outcomes framework and structural causal models formalize interventions and counterfactuals, while causal representation learning extends this view to learned variables and mechanisms~\cite{rubin1974estimating,pearl2009causality,scholkopf2021causalrepresentation}. Work on language models has further examined whether causal information expressed in text can support associational, interventional, and counterfactual reasoning~\cite{kiciman2024causal,jin2023cladder}. Retrieval research has started to use this structure: Causal Graph RAG retrieves causal graphs as structured context for language-model reasoning, and CausalRAG incorporates causal graph construction and tracing into RAG to improve contextual continuity, retrieval precision, and interpretability~\cite{samarajeewa2024causalgraphrag,wang2025causalrag}. These studies show that retrieval can benefit from directional and explanatory structure, but they mainly target knowledge corpora and question answering. In contrast, large procedural memories require relation assessment over executable or reusable skills, where an edge should indicate whether one procedure supports, orders, verifies, or repairs another. Existing graph-based tool and skill retrieval still relies mostly on transitions, dependency labels, similarity, task states, and precondition-effect relations~\cite{liu2024toolnet,liu2026graph,xia2026grasp,chen2026task}. CaSKG addresses this gap by using counterfactual edge evidence to calibrate which candidate skill relations are allowed to shape graph-based retrieval.

\section{Method}\label{sec:method}

\subsection{Framework Overview}\label{subsec:method_overview}
CaSKG is an offline graph-construction and retrieval framework for providing compact procedural context from a reusable skill library. Given a skill library $\mathcal{S}=\{s_1,\ldots,s_n\}$ of $n$ skills, CaSKG constructs a directed candidate graph over ordered skill pairs, calibrates selected edges with direction-conditioned textual counterfactual evidence, and publishes a state-filtered weighted graph. At inference time, CaSKG uses a task query $q$ to identify seed skills, propagates their relevance over the published graph, and retrieves a compact skill bundle for the downstream agent. The agent policy, task interface, and environment interaction loop remain unchanged.

The framework separates relation discovery, edge assessment, graph publication, and task-time retrieval. Candidate induction prioritizes coverage, bringing plausible prerequisite, workflow, and recovery relations into the candidate graph. Counterfactual edge probing estimates whether a candidate relation reflects an operational dependency rather than only topical similarity or unordered co-occurrence. Graph publication determines which relations can transmit query relevance and with what strength. The retrieval stage then expands from task-relevant seeds over the calibrated graph. This design keeps procedural coverage broad while limiting the influence of weak associations.
Figure~\ref{fig:caskg-pipeline} summarizes the four stages of CaSKG: candidate skill graph induction, counterfactual edge probing, edge-state assignment and graph publication, and task-conditioned skill retrieval. The following subsections describe these stages in the same order.

\subsection{Candidate Skill Graph Induction}\label{subsec:relation_construction}

CaSKG first induces a sparse set of directed candidate relations $C\subseteq\mathcal{S}\times\mathcal{S}$. Each $(s_i,s_j)\in C$ denotes the ordered hypothesis $s_i\!\rightarrow\!s_j$, where the source skill $s_i$ may provide operational support for the target skill $s_j$. This stage prioritizes coverage: it collects plausible relations for later calibration while leaving reliability to subsequent assessment.

\paragraph{Multi-source evidence.}
CaSKG is organized around heterogeneous evidence channels from skill descriptions, semantic representations, input/output interfaces, and workflow roles. Lexical and semantic signals capture textual and conceptual affinity between skills, while input/output and structural signals capture interface compatibility and workflow position. Repair evidence, when available, indicates whether one skill can help recover from or complete another procedure. The framework also defines two extension channels for self-evolving skill libraries: future execution logs can contribute trace co-occurrence evidence, and previously recorded relations can preserve continuity across graph updates. The static construction in this paper instantiates the active skill-level channels above, while the trace and relation-history channels remain reserved interfaces for future self-evolving updates. The union of the active recall channels, followed by deduplication and local truncation of low-priority neighbors, yields the candidate set $C$. For edges already in $C$, repair evidence and an optional LLM judge further refine the association score~\cite{zheng2023llmjudge}.

\paragraph{Initial association score.}
For each candidate edge $(s_i,s_j)\in C$, let $\mathcal{A}_{ij}$ be the nonempty set of active scoring signals for the pair, with positive total signal weight. Each $\phi_k(i,j)\in[0,1]$ is the normalized value of signal $k$, and $\lambda_k\geq0$ is the corresponding signal weight. The value $\phi_{\mathrm{struct}}(i,j)$ denotes the normalized structural signal for the ordered pair. The parameters $\tau_{\mathrm{str}}$ and $\eta_{\mathrm{str}}$ denote the activation threshold and retention coefficient for this structural signal. CaSKG first computes the weighted active-signal support $\widetilde{A}_{ij}$ and then applies a structural floor to obtain the initial association score $A_{ij}$:
\begin{equation}
  \begin{aligned}
  \widetilde{A}_{ij}
  &=
  \operatorname{clip}_{[0,1]}\!\left(
  \frac{\sum_{k\in\mathcal{A}_{ij}}\lambda_k\phi_k(i,j)}
       {\sum_{k\in\mathcal{A}_{ij}}\lambda_k}\right),\\
  A_{ij}
  &=
  \begin{cases}
  \max\!\left(\widetilde{A}_{ij},
  \eta_{\mathrm{str}}\phi_{\mathrm{struct}}(i,j)\right),
  & \phi_{\mathrm{struct}}(i,j)>\tau_{\mathrm{str}},\\
  \widetilde{A}_{ij}, & \text{otherwise.}
  \end{cases}
  \end{aligned}
  \label{eq:association_score}
\end{equation}
For any interval $[a,b]$, the operator $\operatorname{clip}_{[a,b]}$ truncates its input to that interval; here, $\operatorname{clip}_{[0,1]}$ bounds the score to the unit interval. The weighted average uses only active signals, so unavailable signals do not become negative evidence. The structural floor preserves strong workflow evidence when $\phi_{\mathrm{struct}}(i,j)$ exceeds $\tau_{\mathrm{str}}$. Thus, $A_{ij}\in[0,1]$ is the initial edge weight on the weighted candidate graph $(C,A)$; pairs outside $C$ have no candidate edge.

The association score serves two purposes. It ranks candidate edges for the limited counterfactual validation pass, and it remains available as discovery support during graph publication. CaSKG selects a budgeted validation frontier $F\subseteq C$ from the candidate graph. Edges in $F$ are sent to the counterfactual probes, while edges in $C\setminus F$ remain unvalidated rather than being treated as negative evidence.

\subsection{Counterfactual Edge Probing}\label{subsec:counterfactual_edge_probing}

An association score can indicate that two skills are related, but it cannot by itself determine whether the ordered edge $s_i\!\rightarrow\!s_j$ captures an operational dependency. A high-scoring association may still connect skills that are interchangeable alternatives, unordered neighbors, or only topically similar. CaSKG therefore probes each selected candidate edge in $F$ with textual counterfactual tests conditioned on the proposed source-to-target direction.

\paragraph{Direction-consistent probes.}
Building on prior studies of language-model counterfactual reasoning and evaluation~\cite{li2023counterfactual,nguyen2024counterfactuals}, CaSKG applies three complementary probes to the skill descriptions of each $(s_i,s_j)\in F$. Let $\mathcal{P}_{m}(s_i,s_j)$ denote the modified relation context for probe type $m$. The removal probe makes the source skill unavailable and tests whether the target skill is impaired, measuring necessity. The substitution probe replaces the source with a low-overlap skill $\widetilde{s}_i$ and tests whether the target skill degrades, measuring source specificity. The reordering probe reverses the proposed relation and tests whether the workflow remains coherent, measuring directionality:
\begin{equation}
  \begin{aligned}
  \mathcal{P}_{\mathrm{rem}}(s_i,s_j)
    &=(\varnothing,s_j),\\
  \mathcal{P}_{\mathrm{sub}}(s_i,s_j)
    &=(\widetilde{s}_i,s_j),\\
  \mathcal{P}_{\mathrm{ord}}(s_i,s_j)
    &=(s_j,s_i),
  \end{aligned}
  \label{eq:counterfactual_probes}
\end{equation}
where $\varnothing$ denotes removing the source skill from the relation context and $\widetilde{s}_i$ denotes the substitute skill used in the substitution test.

For each probe type $m\in\mathcal{M}=\{\mathrm{rem},\mathrm{sub},\mathrm{ord}\}$, the LLM returns an oriented counterfactual support score $e_{ij}^{(m)}\in[0,1]$. A larger score means that removing the source, replacing it with the substitute, or reversing the proposed order would more strongly undermine the hypothesized dependency. The three probe scores share the same support direction and can therefore be aggregated directly: $e_{ij}^{(\mathrm{rem})}$ captures impairment under source removal, $e_{ij}^{(\mathrm{sub})}$ captures degradation under substitution, and $e_{ij}^{(\mathrm{ord})}$ captures loss of workflow coherence under reversal. These scores provide text-level evidence for necessity, specificity, and order dependence of the candidate edge.

\subsection{Bayesian Edge Calibration and Graph Publication}\label{subsec:edge_calibration_publication}

The probe scores are converted into a smoothed relation-reliability estimate rather than being used directly as edge weights. CaSKG then maps each candidate edge to a publication state, which determines whether the edge is published, attenuated, removed, or retained as a low-weight scaffold for coverage.

\paragraph{Reliability estimation.}
CaSKG uses a Beta-form accumulator initialized at $\operatorname{Beta}(1,1)$ to combine the three counterfactual views. For a probe score $e_{ij}^{(m)}$, the binary variable $z_{ij}^{(m)}$ records whether the probe supports the directed relation, and $\delta_{ij}^{(m)}$ records the evidence mass. The parameter $\epsilon_e>0$ is a minimum evidence-mass floor that prevents near-midpoint probe judgments from being ignored entirely:
\begin{equation}
\begin{aligned}
      z_{ij}^{(m)}
  &=\mathbb{I}\!\left[e_{ij}^{(m)}>0.5\right], \\
  \delta_{ij}^{(m)}
  &=\max\!\left(2\left|e_{ij}^{(m)}-0.5\right|,\epsilon_e\right).
\end{aligned}
\label{eq:probe_evidence}
\end{equation}
Here, $\mathbb{I}[\cdot]$ is the indicator function. The midpoint $0.5$ determines evidence polarity, while the distance from the midpoint determines how much evidence the probe contributes. The aggregated Beta parameters are
\begin{equation}
  \begin{aligned}
  \alpha_{ij}
  &=1+\sum_{m\in\mathcal{M}}
    z_{ij}^{(m)}\delta_{ij}^{(m)},\\
  \beta_{ij}
  &=1+\sum_{m\in\mathcal{M}}
    \left(1-z_{ij}^{(m)}\right)\delta_{ij}^{(m)}.
  \end{aligned}
  \label{eq:beta_update}
\end{equation}
where $\alpha_{ij}$ accumulates positive evidence for the directed relation and $\beta_{ij}$ accumulates evidence against it. The normalized Beta mean gives the smoothed relation-reliability score:
\begin{equation}
  c_{ij}=\frac{\alpha_{ij}}{\alpha_{ij}+\beta_{ij}}.
  \label{eq:beta_mean}
\end{equation}
The score $c_{ij}\in[0,1]$ summarizes probe-derived support for the ordered relation $s_i\!\rightarrow\!s_j$.

\paragraph{State-gated publication.}
The association score $A_{ij}$ represents discovery support, and the reliability score $c_{ij}$ represents counterfactual support. CaSKG uses a symmetric confirmation threshold $\tau_c\in(0.5,1)$ to assign an assessment state $\sigma_{ij}$:
\begin{equation}
  \sigma_{ij}=
  \begin{cases}
  \mathrm{confirmed},
  &(s_i,s_j)\in F\ \land\ c_{ij}>\tau_c,\\
  \mathrm{rejected},
  &(s_i,s_j)\in F\ \land\ c_{ij}<1-\tau_c,\\
  \mathrm{uncertain},
  &(s_i,s_j)\in F\ \land\ 1-\tau_c\leq c_{ij}\leq\tau_c,\\
  \mathrm{unvalidated},
  &(s_i,s_j)\in C\setminus F,
  \end{cases}
  \qquad (s_i,s_j)\in C.
  \label{eq:relation_state}
\end{equation}
Confirmed edges receive strong evidence from the probes, rejected edges receive counter-evidence, uncertain edges remain within the middle band, and unvalidated edges are candidates that did not enter the validation frontier. A bounded subset $E_{\mathrm{scaf}}\subseteq C\setminus F$ is retained as scaffold edges for coverage; their weights are governed by the scaffold attenuation rather than by probe-derived reliability.

Define $\widehat{c}_{ij}=c_{ij}$ for $(s_i,s_j)\in F$ and $\widehat{c}_{ij}=0$ otherwise. Let $\epsilon_w>0$ be the floor for positive published edge weights. Let $\rho_{\mathrm{unc}}$ and $\rho_{\mathrm{scaf}}$ be the attenuation coefficients for uncertain and scaffold edges, with $1>\rho_{\mathrm{unc}}>\rho_{\mathrm{scaf}}>0$. CaSKG combines discovery support and probe-derived reliability into $b_{ij}$ and then applies the state-dependent publication gate $\rho_{ij}$:
\begin{equation}
  \begin{aligned}
  b_{ij}
  &=\max\!\left(A_{ij},\widehat{c}_{ij},\epsilon_w\right),\\
  \rho_{ij}
  &=
  \begin{cases}
  1, & \sigma_{ij}=\mathrm{confirmed},\\
  \rho_{\mathrm{unc}}, & \sigma_{ij}=\mathrm{uncertain},\\
  \rho_{\mathrm{scaf}}, &
  \sigma_{ij}=\mathrm{unvalidated}
  \ \land\ (s_i,s_j)\in E_{\mathrm{scaf}},\\
  0, & \text{otherwise,}
  \end{cases}\\
  w^{\mathrm{pub}}_{ij}
  &=
  \begin{cases}
  \operatorname{clip}_{[\epsilon_w,1]}(\rho_{ij}b_{ij}),
  &\rho_{ij}>0,\\
  0,&\rho_{ij}=0.
  \end{cases}
  \end{aligned}
  \label{eq:publication_weight}
\end{equation}
Here, $b_{ij}$ keeps the stronger available support source, $\rho_{ij}$ controls how the edge state affects propagation, and $w^{\mathrm{pub}}_{ij}$ is the final published edge weight. Confirmed relations retain full support, uncertain relations are downweighted, rejected relations are removed, and selected unvalidated candidates serve as attenuated scaffold links.

The publication step produces the graph
\begin{equation}
  \begin{aligned}
      G_{\mathrm{pub}}
      &=(\mathcal{S},E_{\mathrm{pub}},W,\Sigma),\\
      E_{\mathrm{pub}}
      &=\{(s_i,s_j)\in C:w^{\mathrm{pub}}_{ij}>0\},
  \end{aligned}
  \label{eq:published_graph}
\end{equation}
where $E_{\mathrm{pub}}$ is the published edge set, $W$ contains the published weights, and $\Sigma$ contains the assessment states. This graph is the frozen structure used by the retrieval stage.

\subsection{Task-Conditioned Skill Retrieval}\label{subsec:graph_retrieval}

At inference time, CaSKG uses $G_{\mathrm{pub}}$ as the structural substrate for skill retrieval. Given a task query $q$, lexical and semantic rankings provide initial task-relevant seeds. These seeds are reranked and converted into an inverse-rank-weighted seed distribution $\pi_q$, where higher-ranked skills receive larger probability mass. In the personalized PageRank update below, $\pi_q$ anchors the restart term to the current task, while graph propagation allows related skills to enter the retrieved context even when they are not direct semantic neighbors of the query.

\paragraph{Query-conditioned diffusion.}
Let $T$ be the row-normalized transition matrix derived from the published graph $G_{\mathrm{pub}}$ and its edge weights $W$. Let $\gamma\in(0,1)$ be the restart coefficient, which controls how strongly each update returns to $\pi_q$, and let $p^{(t)}$ be the skill relevance distribution after $t$ propagation steps. Starting from $p^{(0)}=\pi_q$, personalized PageRank~\cite{haveliwala2002topicsensitive} iterates
\begin{equation}
  p^{(t+1)}=\gamma\pi_q+(1-\gamma)T^{\top}p^{(t)}.
  \label{eq:task_diffusion}
\end{equation}
After convergence, writing $p$ for the limiting relevance distribution, CaSKG ranks skills by $p$ and returns the highest-ranked skill summaries or procedures as the task context. The restart term keeps the expansion tied to the query, while the state-gated edge weights determine which procedural relations transmit relevance and how strongly they do so. In this way, CaSKG couples task-local semantic seeds with reliability-weighted relational expansion, allowing procedurally related skills to enter the context while reducing the influence of weak relations.

\section{Experiments}
\label{sec:experiments}

The experiments examine whether calibrating skill-graph edges before retrieval improves both task outcomes and interaction behavior. The comparison is designed around three questions: whether CaSKG improves over full-library access, independent semantic retrieval, and an existing graph-retrieval baseline; whether any task-score gain comes with additional environment interaction cost; and whether the observed gains can be explained by more coherent procedural skill bundles. We evaluate all methods under a frozen Skill1000 library on two interactive benchmarks, report the common protocol and main results, and then analyze task-type and trajectory-level behavior.

\subsection{Experimental Setup}
\label{sec:experimental-setup}

\paragraph{Benchmarks.}
We use two interactive benchmarks with complete evaluated cohorts. ALFWorld ID-140 contains 140 in-distribution household-task episodes and measures whether the agent completes the goal. ScienceWorld U211 contains 211 evaluated science-task episodes and reports the official best score for each episode. The main comparison uses a Skill1000 library. For every method, the library or retrieval structure is built offline and kept fixed during evaluation, so performance differences reflect how skill context is exposed rather than online adaptation of the skill store.

\paragraph{Baselines.}
We compare four retrieval settings that correspond to different ways of managing the tradeoff between coverage and noise. Vanilla Skills exposes the complete Skill1000 catalog without graph retrieval, maximizing recall but leaving filtering to the agent. Vector Skills retrieves skills independently by embedding similarity, reducing context size but ignoring procedural links among skills. GoS uses dependency-aware graph construction and graph propagation to recover skills connected to the query. CaSKG uses the counterfactual-causal, state-weighted graph constructed by the procedure in Section~\ref{sec:method}.

\paragraph{Models and evaluation.}
The main comparison covers MiniMax-M2.7~\cite{chen2026minimax}, GLM-5.2~\cite{zeng2026glm}, Kimi-K2.6~\cite{kimi2026k2.6}, Qwen3.5-397B-A17B~\cite{qwen35blog}, DeepSeek-V4-Flash~\cite{xu2026deepseek}, and GPT-5.6-Luna~\cite{openai2026gpt56changelog}. Within each benchmark, methods use the same task cohort, prompt, evaluator, episode limits, and environment interaction loop. CaSKG publishes its graph before evaluation and does not update the graph or add a separate online planner during an episode. Each accepted record stores per-episode outcomes, environment steps, retrieved context, and infrastructure diagnostics.

\paragraph{Metrics and reporting rules.}
On ALFWorld, $R$ is the success rate over 140 tasks and is reported as a percentage. On ScienceWorld, $R$ is the mean best official score over 211 episodes and is not a percentage. We also report \emph{Steps}, the arithmetic mean number of environment interactions per episode over the complete valid cohort; higher $R$ and fewer steps are better. Steps measures observed interaction consumption during task execution rather than token usage, wall-clock latency, retrieval latency, or graph-construction cost.

\subsection{Main Results}
\label{subsec:main-results-experiments}

Table~\ref{tab:skill1000-results} shows that CaSKG provides the strongest overall task-performance--interaction-cost profile. It obtains the highest task score for every backbone on both benchmarks and uses fewer mean steps than GoS in all twelve model--benchmark settings. Averaged across the six backbones, the ScienceWorld score rises from 72.62 with GoS to 80.50 with CaSKG, while ALFWorld success rises from 80.01\% to 86.79\%. This consistency across weaker and stronger backbones is important: the gain is not confined to a single model family or to cases where the base model is weak. Instead, the results indicate that improving the retrieval graph changes the usefulness of the context supplied to the same downstream agent loop.

\begin{table}[t]
\caption{Reward and mean interaction steps at Skill1000 on ALFWorld ID-140 and ScienceWorld U211.}
\label{tab:skill1000-results}
\centering
\fontsize{6.2}{7.0}\selectfont
\setlength{\tabcolsep}{1.0pt}
\renewcommand{\arraystretch}{1.02}
\begin{tabularx}{\columnwidth}{@{}>{\hsize=1.65\hsize\raggedright\arraybackslash}X>{\hsize=.75\hsize\raggedright\arraybackslash}X*{4}{>{\hsize=.90\hsize\centering\arraybackslash}X}@{}}
\toprule
Model & Method &
\multicolumn{2}{c}{ALFWorld ID-140} &
\multicolumn{2}{c}{ScienceWorld U211} \\
\cmidrule(lr){3-4}\cmidrule(lr){5-6}
 & & $R\,(\%)\uparrow$ & Steps$\downarrow$ & $R\uparrow$ & Steps$\downarrow$ \\
\midrule
\multirow[c]{4}{*}{MiniMax-M2.7}
 & Vanilla & 42.90 & 22.54 & 45.90 & 21.73 \\
 & Vector & 45.70 & 22.84 & 43.21 & 21.45 \\
 & GoS & \secondresult{63.60} & \secondresult{19.69} & \secondresult{55.85} & \secondresult{18.91} \\
\rowcolor{caskgblue}
 & CaSKG & \bestresult{73.57} & \bestresult{18.44} & \bestresult{68.33} & \bestresult{17.45} \\
\midrule
\multirow[c]{4}{*}{GLM-5.2}
 & Vanilla & 95.00 & 11.05 & 75.50 & 17.03 \\
 & Vector & \secondresult{96.43} & 10.12 & 77.07 & 16.65 \\
 & GoS & 95.71 & \secondresult{9.91} & \secondresult{80.33} & \secondresult{15.75} \\
\rowcolor{caskgblue}
 & CaSKG & \bestresult{97.86} & \bestresult{9.69} & \bestresult{85.11} & \bestresult{14.52} \\
\midrule
\multirow[c]{4}{*}{Kimi-K2.6}
 & Vanilla & 77.90 & 16.07 & 72.23 & 18.91 \\
 & Vector & 90.00 & 13.49 & 72.58 & 17.55 \\
 & GoS & \secondresult{93.60} & \secondresult{13.08} & \secondresult{76.82} & \secondresult{16.15} \\
\rowcolor{caskgblue}
 & CaSKG & \bestresult{95.00} & \bestresult{12.34} & \bestresult{83.88} & \bestresult{15.43} \\
\midrule
\multirow[c]{4}{*}{\shortstack[l]{Qwen3.5-397B-A17B}}
 & Vanilla & 79.30 & 15.60 & \secondresult{63.72} & 18.34 \\
 & Vector & 78.60 & 15.49 & 62.60 & 18.51 \\
 & GoS & \secondresult{88.60} & \secondresult{14.15} & 63.18 & \secondresult{17.08} \\
\rowcolor{caskgblue}
 & CaSKG & \bestresult{92.14} & \bestresult{11.60} & \bestresult{74.97} & \bestresult{15.56} \\
\midrule
\multirow[c]{4}{*}{\shortstack[l]{DeepSeek-V4-Flash}}
 & Vanilla & 72.86 & 16.91 & 64.84 & 18.49 \\
 & Vector & \secondresult{78.57} & \secondresult{16.89} & 68.65 & 18.39 \\
 & GoS & 77.86 & 17.09 & \secondresult{73.45} & \secondresult{16.20} \\
\rowcolor{caskgblue}
 & CaSKG & \bestresult{86.43} & \bestresult{14.41} & \bestresult{83.40} & \bestresult{15.61} \\
\midrule
\multirow[c]{4}{*}{\shortstack[l]{GPT-5.6-Luna}}
 & Vanilla & \secondresult{72.86} & \bestresult{17.74} & 84.09 & 14.99 \\
 & Vector & 55.00 & 22.06 & 84.09 & 14.40 \\
 & GoS & 60.71 & 21.86 & \secondresult{86.08} & \secondresult{14.22} \\
\rowcolor{caskgblue}
 & CaSKG & \bestresult{75.71} & \secondresult{17.79} & \bestresult{87.33} & \bestresult{13.18} \\
\bottomrule
\end{tabularx}
\vspace{1.5pt}

\parbox{\columnwidth}{\fontsize{5.55}{6.3}\selectfont
\textbf{Bold} and \underline{underlined} values are the best and second-best reported results within each model and metric; higher $R$ and lower Steps are better. Rankings require at least two reported values. CaSKG rows are shaded blue.}
\end{table}

\paragraph{Overall comparison.}
The four retrieval settings expose different failure modes. \textsc{Vanilla Skills} keeps the entire library available, so it can include useful procedures but also forces the model to search through irrelevant alternatives during task execution. \textsc{Vector Skills} reduces this burden, but it retrieves each skill as an independent text item and therefore misses procedures whose value comes from workflow position rather than lexical similarity. \textsc{GoS} improves over vector retrieval by allowing relevance to move through a skill graph, which is especially helpful when the required skill bundle contains prerequisites or follow-up actions. CaSKG keeps this relational benefit while filtering the graph through counterfactual edge evidence. The result is a retrieval context that is neither an unfiltered catalog nor a set of isolated nearest neighbors; it is a compact bundle shaped by validated procedural relations.

\paragraph{ScienceWorld.}
ScienceWorld is the benchmark where the benefit of calibrated graph retrieval is most visible. Many tasks require a sequence of preparation, operation, observation, and classification steps rather than a single action named in the task instruction. Under this condition, direct semantic matches often recover only part of the procedure. GoS already improves over Vector Skills for all six backbones, confirming that graph propagation helps retrieve supporting skills beyond the nearest textual neighbors. CaSKG adds a second layer of improvement by making the propagated edges more selective. The largest gaps appear for MiniMax-M2.7 and DeepSeek-V4-Flash, where CaSKG improves over GoS by 12.48 and 9.95 points, respectively. The advantage narrows for stronger backbones, but it remains positive: CaSKG is still ahead of GoS by 4.78 points for GLM-5.2 and 1.25 points for GPT-5.6-Luna. This pattern suggests that CaSKG is most useful when a model needs retrieval to supply missing procedural structure, while still providing incremental benefit when the model already solves many episodes.

\paragraph{ALFWorld.}
ALFWorld shows the same advantage in a more stateful household-control setting. Success requires not only identifying the target object or receptacle, but also preserving the order of navigation, pickup, state change, and placement. The largest separation again occurs for MiniMax-M2.7: CaSKG improves on GoS by 9.97 percentage points and on both non-graph settings by more than 27 points. Kimi-K2.6 and Qwen3.5-397B-A17B follow the same ranking pattern, while GLM-5.2 is already near the success ceiling and leaves less room for improvement. GPT-5.6-Luna reveals why retrieval design still matters for strong models: Vector Skills and GoS fall sharply below Vanilla Skills, whereas CaSKG recovers the best success rate among all settings. The result indicates that structural retrieval is beneficial only when the structure suppresses misleading links; otherwise, graph propagation can be worse than exposing the full catalog.
\begin{figure}[t]
    \centering
    \includegraphics[width=\columnwidth]{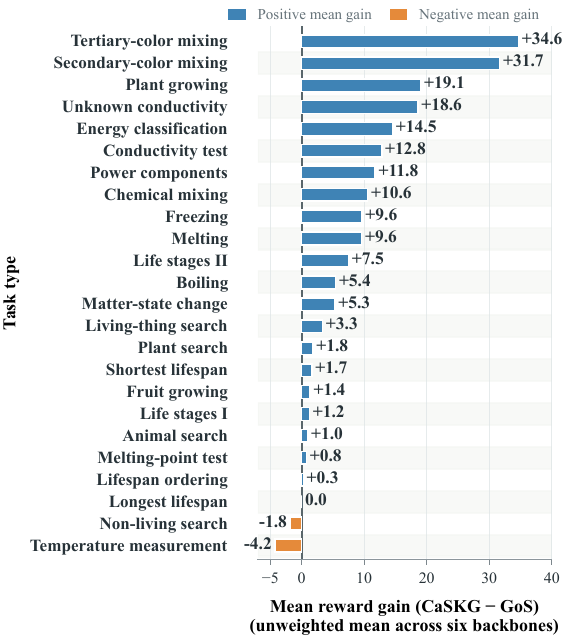}
    \caption{Task-type reward gains of CaSKG over GoS on ScienceWorld U211. All 24 task types are shown individually and ordered by gain; each bar is the unweighted mean reward difference across the six evaluated backbones. Blue and orange indicate positive and negative gains, respectively.}
    \Description{Horizontal bar chart of the mean CaSKG score difference over GoS
across 24 ScienceWorld task types. CaSKG improves on 21 types, ties on one,
and trails on non-living-thing search and temperature measurement.}
    \label{fig:task-type-gain-bar}
\end{figure}

\paragraph{Environment-interaction cost.}
The task-score gains are not obtained by spending more environment interactions. Under the shared episode limit, CaSKG reduces the ScienceWorld six-model mean to 15.29 steps, compared with 16.39 for GoS and roughly 18 steps for the two non-graph settings. It is the shortest method for every ScienceWorld backbone. On ALFWorld, CaSKG averages 14.05 steps, below GoS at 15.96 and the two non-graph settings at roughly 16.7--16.8 steps. It also uses the fewest steps in five of six ALFWorld backbones; the only exception is GPT-5.6-Luna, where Vanilla Skills is marginally shorter (17.74 versus 17.79) but less successful. Across both benchmarks, CaSKG uses fewer steps than GoS in all twelve model--benchmark settings and has the lowest mean step count among all methods in eleven. The more plausible interpretation is that calibrated graph retrieval reduces exploratory and corrective actions by giving the agent a more executable procedure, rather than simply allowing it to interact longer. This is still an aggregate interaction-cost pattern, not a claim about token usage, latency, graph-construction cost, or success-conditioned efficiency.

Overall, the results support the intended advantage of CaSKG: it improves the structure of retrieved context rather than merely changing the amount of context. The following qualitative analysis examines this explanation at the task-type and trajectory levels.

\subsection{Qualitative Analysis}
\label{subsec:qualitative-main-results}

We use qualitative analysis to connect the aggregate improvements in Table~\ref{tab:skill1000-results} to the retrieval mechanism. The first view is a task-type breakdown on ScienceWorld. Figure~\ref{fig:task-type-gain-bar} reports the unweighted mean gain of CaSKG over GoS across the six evaluated backbones for all 24 U211 task types, ordered by gain. CaSKG improves on 21 task types, ties on one, and trails GoS on two. The largest gains appear in tertiary- and secondary-color mixing, plant growing, unknown conductivity, and energy classification. These categories typically require multiple dependent operations: setting up materials, applying a transformation, observing the result, and mapping the observation to a final answer. The two negative cases, non-living-thing search and temperature measurement, are more direct retrieval or measurement tasks, where additional graph expansion can offer less advantage and may occasionally introduce distraction.

The task-type pattern suggests that CaSKG is most useful when success depends on preserving a chain of operations rather than retrieving one obviously named skill. We therefore examine one representative task from each benchmark and compare all four methods. The examples are not intended as independent proof of the aggregate results; instead, they illustrate the failure modes behind the table: full-library exposure can leave the agent to choose among noisy alternatives, vector retrieval can miss non-lexical dependencies, and graph retrieval can still be misled if weak edges carry relevance into the retrieved context.

\paragraph{ScienceWorld: conductivity testing.}
The conductivity example shows how CaSKG turns a partially matched task into an executable procedure. Determining whether sodium chloride conducts electricity requires focusing on the material, assembling a valid circuit, observing the result, classifying the material, and placing it in the corresponding box. In a MiniMax-M2.7 episode of \texttt{test-conductivity}, CaSKG retrieved guidance covering conductivity testing, circuit construction and connection, material classification, and final placement. The agent completed the circuit, identified sodium chloride as nonconductive, and placed it in the green box, scoring 100 in 24 steps. The baselines fail in different ways. GoS reached 55 after 30 steps because unrelated utilities entered the context and the agent spent its remaining interactions repairing connection commands. Vanilla Skills also scored 55 in 29 steps; although the full catalog contained relevant skills, it did not prevent repeated trials of alternative endpoint descriptions. Vector Skills scored 5 in 30 steps because semantic retrieval supplied mostly unrelated technical utilities and no workable circuit plan. The advantage of CaSKG in this case is not simply that it retrieved a skill about conductivity. It retrieved a bundle that preserved the preparation--operation--verification--placement dependency chain needed to finish the task.

\paragraph{ALFWorld: cooling and placement.}
The ALFWorld example highlights the same mechanism in a stateful household workflow. Cooling an apple and placing it on a countertop requires the agent to search, pick up the object, locate or use the cooling appliance, track the changed object state, and complete the final placement. In the GLM-5.2 episode \texttt{pick\_cool\_then\_place\_in\_recep-\allowbreak{}Apple-\allowbreak{}None-\allowbreak{}CounterTop-\allowbreak{}14}, CaSKG retrieved complementary guidance for locating the appliance, cooling the object, tracking its state, and operating the destination receptacle. It found the apple, cooled it, and placed it on the countertop in 27 steps. Vanilla Skills, Vector Skills, and GoS all received zero reward at the 30-step limit, but their failures differ. Vanilla Skills reached cooling but did not complete the final move, showing that recall alone does not guarantee procedural completion. Vector Skills retrieved mostly unrelated simulation and parallelization skills, so its compact context omitted key household dependencies. GoS found the apple late, completed cooling, and exhausted its budget before placement despite retrieving a temperature-regulation skill. The case illustrates why relevance to a state-changing action is insufficient: successful retrieval must also preserve the search, object-state tracking, and final placement relations that define the task.

The two examples lead to the same conclusion as the aggregate results. CaSKG succeeds not because it names the target operation more often, but because it returns a usable chain of adjacent skills: preparation, state-changing action, verification, and final completion. The baselines expose the complementary failures behind this result. Full-library access has recall but weak ordering pressure, vector retrieval is compact but can omit non-lexical dependencies, and uncalibrated graph propagation can include relevant-looking edges that do not support the next executable step. This explains why CaSKG is most advantageous when the task objective hides several dependent subgoals. At the same time, the negative task types in Figure~\ref{fig:task-type-gain-bar} show that graph structure is not universally beneficial; when the task is close to a direct search or measurement operation, the additional relational context has less room to help. The main result of the qualitative analysis is therefore selective procedural expansion: CaSKG keeps retrieval broad enough to include prerequisites and follow-up actions, but constrains propagation so that weak associations are less likely to dominate the agent's context.

\section{Ablation Study}\label{sec:ablation-study}

The ablation study further attributes the gains in Section~\ref{sec:experiments} to the main design choices in CaSKG. We organize the analysis around two diagnostic questions. The scale study examines whether calibrated graph retrieval remains effective as the skill library becomes larger and more diverse. The component study examines how broad candidate induction, judge-assisted scoring, and counterfactual correction with state-gated publication contribute to the final retrieval graph. Throughout this section, $R$ denotes ALFWorld success rate and \emph{Steps} denotes the mean number of environment interactions; higher $R$ and fewer steps are better.

\subsection{Sensitivity to Skill Library Size}\label{subsec:skill-library-size-ablation}

Larger skill libraries provide richer procedural coverage and create a stronger test of relation calibration, since retrieval must select useful procedural neighborhoods from a broader set of candidate relations. We therefore use the ALFWorld ID-140 scale records to compare CaSKG with GoS at 200, 500, 1,000, and 2,000 skills. The comparison covers MiniMax-M2.7 and Qwen3.5-397B-A17B. For MiniMax-M2.7, CaSKG assesses 500 candidate relations at 200--1,000 skills and expands the validation frontier to 2,000 relations at the 2,000-skill scale, matching the larger construction setting used at that scale.

\begin{table}[t]
\caption{Archived skill-library scale comparison on ALFWorld ID-140 for MiniMax-M2.7 and Qwen3.5-397B-A17B.}
\label{tab:skill-library-size-ablation}
\centering
\fontsize{6.65}{7.45}\selectfont
\setlength{\tabcolsep}{1.15pt}
\renewcommand{\arraystretch}{1.03}
\begin{tabular*}{\columnwidth}{@{\extracolsep{\fill}}ll*{5}{r}@{}}
\toprule
Model & Skills & \multicolumn{3}{c}{$R\,(\%)\uparrow$} & \multicolumn{2}{c}{Steps$\downarrow$} \\
\cmidrule(lr){3-5}\cmidrule(lr){6-7}
 &  & CaSKG & GoS & $\Delta R$ & CaSKG & GoS \\
\midrule
\multirow{4}{*}{MiniMax-M2.7}
 & 200   & \textbf{57.14} & 50.00 & +7.14  & \textbf{20.73} & 22.21 \\
 & 500   & \textbf{67.86} & 45.00 & +22.86 & \textbf{19.95} & 23.07 \\
 & 1,000 & \textbf{73.57} & 63.60 & +9.97  & \textbf{18.44} & 19.69 \\
 & 2,000 & \textbf{70.00} & 54.29 & +15.71 & \textbf{18.71} & 21.32 \\
\addlinespace[0.6pt]
\cmidrule(lr){1-7}
\multirow{4}{*}{\shortstack[l]{Qwen3.5-397B-\\A17B}}
 & 200   & \textbf{85.00} & 76.43 & +8.57  & \textbf{14.50} & 16.25 \\
 & 500   & \textbf{94.29} & 72.86 & +21.43 & \textbf{12.48} & 16.94 \\
 & 1,000 & \textbf{92.14} & 88.60 & +3.54  & \textbf{11.60} & 14.15 \\
 & 2,000 & \textbf{91.43} & 77.86 & +13.57 & \textbf{12.31} & 16.47 \\
\bottomrule
\end{tabular*}
\vspace{1.5pt}

\parbox{\columnwidth}{\fontsize{6.35}{7.1}\selectfont
$R$ is success rate, $\Delta R=R_{\mathrm{CaSKG}}-R_{\mathrm{GoS}}$ is measured in percentage points, and Steps is the mean number of environment interactions. Values reproduce archived aggregate records; Qwen GoS values are archived aggregate counterparts rather than episode directories. In the MiniMax runs, CaSKG assesses 500 candidate relations at 200--1,000 skills and 2,000 relations at 2,000 skills, so this is a descriptive system-level scale comparison. Bold denotes the better observed result in each method pair.}
\end{table}

\begin{figure}[t]
    \centering
    \includegraphics[width=0.95\columnwidth]{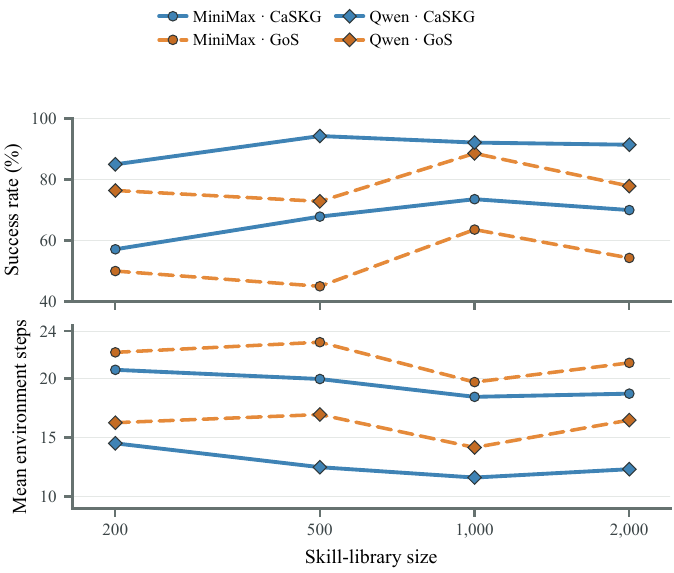}
    \caption{Skill-library-size sensitivity on ALFWorld ID-140 for MiniMax-M2.7 and Qwen3.5-397B-A17B. The upper panel reports task success rate and the lower panel reports mean environment steps for CaSKG and GoS at four library sizes.}
    \Description{Two-panel line chart comparing CaSKG and GoS for MiniMax-M2.7
and Qwen3.5-397B-A17B at skill-library sizes 200, 500, 1,000, and 2,000.
CaSKG has higher success rates and fewer mean environment steps at every size.}
    \label{fig:library-size-sensitivity}
\end{figure}

Figure~\ref{fig:library-size-sensitivity} visualizes the same scale comparison and highlights CaSKG's stable advantage under different library sizes. On MiniMax-M2.7, CaSKG improves success over GoS at every scale, with gains of 7.14, 22.86, 9.97, and 15.71 percentage points from 200 to 2,000 skills. On Qwen3.5-397B-A17B, the corresponding gains are 8.57, 21.43, 3.54, and 13.57 points. The improvement is largest at 500 skills for both backbones, and CaSKG still adds 3.54 points at the strong 1,000-skill Qwen setting where GoS already reaches 88.60\% success. This pattern indicates that CaSKG benefits from calibrated procedural structure across both moderate and large libraries. Its relative advantage is most visible when the graph contains rich procedural neighbors and the retrieval process can use edge calibration to focus propagation.

The step results provide a second view of the same mechanism. CaSKG uses fewer mean environment interactions than GoS at every tested scale for both backbones. For MiniMax-M2.7, the reductions are 1.48, 3.12, 1.25, and 2.61 steps; for Qwen3.5-397B-A17B, they are 1.75, 4.46, 2.55, and 4.16 steps. The success gains are therefore accompanied by lower interaction cost. This pattern is consistent with the role of the calibrated graph: it gives the agent a more focused procedural neighborhood, reducing exploratory and corrective actions after retrieval. MiniMax reaches its best observed CaSKG success at 1,000 skills, while Qwen reaches its best observed CaSKG success at 500 skills, showing that the most useful library size can depend on the backbone. Across these settings, CaSKG maintains the stronger success--step profile.

\subsection{Component Analysis of Graph Construction}\label{subsec:graph-construction-ablation}

The component ablation examines the two design pressures behind CaSKG's graph construction. The first stage should recover a sufficiently broad set of candidate relations, including procedural dependencies that are difficult to capture from surface similarity alone. The second stage should select the relations that are reliable enough to influence retrieval. We evaluate four pipeline configurations under the Skill1000 setting with MiniMax-M2.7 on ALFWorld ID-140. All variants use Qwen3-Embedding-8B with 4,096-dimensional embeddings, share the same 140-task cohort, retrieval mode, 30-step limit, and downstream execution protocol. Full CaSKG and the two first-stage ablations each assess 500 candidate relations. The full-candidate variant publishes the complete candidate graph, providing a direct comparison between selective publication and maximum graph density.

\begin{table}[t]
\caption{Component ablation of CaSKG's graph-construction pipeline with MiniMax-M2.7 and the Skill1000 library on ALFWorld ID-140.}
\label{tab:ablation-design}
\centering
\fontsize{7.0}{8.0}\selectfont
\setlength{\tabcolsep}{1.2pt}
\renewcommand{\arraystretch}{1.04}
\begin{tabularx}{\columnwidth}{@{}>{\hsize=1.75\hsize\raggedright\arraybackslash}X*{5}{>{\hsize=.85\hsize\raggedleft\arraybackslash}X}@{}}
\toprule
Variant & $|C|$ & $|F|$ & $|E_{\mathrm{pub}}|$ & $R\,(\%)\uparrow$ & Steps$\downarrow$ \\
\midrule
\rowcolor{caskgblue}
Full CaSKG & 9,937 & 500 & 3,292 & \textbf{73.57} & \textbf{18.44} \\
Semantic-only & 3,982 & 500 & 2,698 & 67.14 & 19.21 \\
w/o LLM judge & 9,753 & 500 & 3,188 & 71.43 & 18.79 \\
Publish all candidates & 9,937 & 0 & 9,937 & 71.43 & 18.74 \\
\bottomrule
\end{tabularx}
\vspace{1.5pt}

\parbox{\columnwidth}{\fontsize{7.0}{8.0}\selectfont
$|C|$, $|F|$, and $|E_{\mathrm{pub}}|$ denote the numbers of candidate, counterfactually assessed, and published relations, respectively. $R$ is success rate (\%), and Steps is the mean number of environment interactions; higher $R$ and lower Steps are better.}
\end{table}

Table~\ref{tab:ablation-design} shows selective publication gives the strongest configuration. Full CaSKG reaches 73.57\% success with 18.44 mean steps while publishing 3,292 of 9,937 candidate relations. The full-candidate variant publishes all 9,937 relations and reaches 71.43\% success with 18.74 mean steps. Because both configurations start from the same candidate set, the comparison highlights the value of counterfactual correction and state-gated publication. The selected graph provides enough relational coverage for retrieval while keeping propagation concentrated on higher-confidence procedural relations.

The semantic-only variant further highlights the role of multi-signal candidate induction. It constructs 3,982 candidate relations and publishes 2,698 relations, reaching 67.14\% success with 19.21 mean steps. Full CaSKG raises success by 6.43 percentage points and lowers the mean step count by 0.77 steps. The improvement shows that non-semantic evidence supplies useful procedural neighbors whose value depends on workflow role, interface compatibility, co-occurrence, or repair evidence beyond lexical similarity alone. This ablation evaluates the multi-signal candidate stage as an integrated module and supports its contribution to downstream execution.

The no-judge variant shows that the optional judge signal provides additional calibration on top of deterministic construction signals. The candidate and published graph sizes remain close to the full configuration, while full CaSKG adds 2.14 percentage points of success and reduces the mean step count from 18.79 to 18.44. This pattern suggests that the judge signal refines borderline relation scores that affect which edges survive publication and how strongly they influence propagation. The comparison treats the judge as part of the complete graph-construction pipeline and shows its incremental contribution under the reported configuration.

The component results give a sharper explanation of the main comparison. CaSKG benefits from both sides of the pipeline: multi-signal construction supplies a candidate graph broad enough to include non-lexical procedural dependencies, and counterfactual state-gated publication turns that broad candidate set into a confidence-weighted retrieval graph. The LLM judge further refines relation scoring under this configuration. Together, the ablations support the design principle of CaSKG: retrieve from a graph that is broad at the candidate stage and selective at the publication stage.
\section{Conclusion}\label{sec:conclusion}
This paper studies skill retrieval for LLM agents as the problem of exposing compact and executable procedural context from a large reusable skill library. CaSKG addresses this problem by separating high-recall association discovery from edge-confidence calibration. It constructs a directed candidate graph from heterogeneous skill signals, evaluates selected relations with direction-conditioned textual counterfactual probes, and publishes a state-filtered weighted graph for task-conditioned retrieval. Across ALFWorld and ScienceWorld, CaSKG achieves the best reported task score in all twelve model--benchmark settings. Relative to GoS, the six-model macro-average score increases from 72.62 to 80.50 on ScienceWorld and from 80.01\% to 86.79\% on ALFWorld, while using fewer observed environment interactions. Task-type,
trajectory, library-scale, and ablation analyses are consistent with calibrated relations helping preserve prerequisites, state-changing actions, verification routines, and completion steps within the retrieved skill bundle.

Overall, these findings suggest that edge-confidence calibration is a useful design principle for scalable agent memory. By combining broad candidate recall with selective graph publication, CaSKG retrieves compact skill context while preserving the operational structure needed to execute complex tasks.

\section*{Ethics and Privacy Statement}

This work studies skill retrieval for LLM agents in simulated ALFWorld and ScienceWorld environments using reusable skill libraries and aggregate task outcomes; it does not collect personal data, involve human subjects, or infer sensitive attributes. The main broader-impact consideration is that more reliable procedural retrieval can make autonomous agents more capable, which benefits reproducible tool use and controlled task execution but should still be deployed with task-appropriate safeguards when connected to external tools or real-world environments. CaSKG is an offline retrieval-graph construction method that does not introduce a new action policy or expand the agent's permissions, and the experiments are conducted within benchmark-defined environments and evaluation protocols.

\bibliographystyle{ACM-Reference-Format}
\bibliography{refs/references}

\end{document}